# An unsupervised latent/output physics-informed convolutional-LSTM network for solving partial differential equations using peridynamic differential operator

Arda Mavi [a†], Ali Can Bekar [a†], Ehsan Haghighat [b], Erdogan Madenci [a,1]

[a]*Department of Aerospace and Mechanical Engineering, The University of Arizona, Tucson, AZ 85721, USA*
[b]*Department of Civil and Environmental Engineering, Massachusetts Institute of Technology, Cambridge, MA 02139, USA*

**Abstract**
This study presents a novel unsupervised convolutional Neural Network (NN) architecture with nonlocal interactions for solving Partial Differential Equations (PDEs). The nonlocal Peridynamic Differential Operator (PDDO) is employed as a convolutional filter for evaluating derivatives the field variable. The NN captures the time-dynamics in smaller latent space through encoder-decoder layers with a Convolutional Long-short Term Memory (ConvLSTM) layer between them. The ConvLSTM architecture is modified by employing a novel activation function to improve the predictive capability of the learning architecture for physics with periodic behavior. The physics is invoked in the form of governing equations at the output of the NN and in the latent (reduced) space. By considering a few benchmark PDEs, we demonstrate the training performance and extrapolation capability of this novel NN architecture by comparing against Physics Informed Neural Networks (PINN) type solvers. It is more capable of extrapolating the solution for future timesteps than the other existing architectures.

## 1. Introduction

Partial Differential Equations (PDEs) are crucial tools to understand and describe phenomena such as sound, diffusion, electrodynamics and fluid dynamics. However, obtaining analytical solutions to most PDEs is challenging and may be not possible. Therefore, variety of numerical techniques for solving different PDEs are developed, including Finite Element Method (FEM), Finite Difference Method (FDM), and Spectral Methods to name a few [1−4]. Although these methods are well established and capable of solving a wide range of PDEs, the use of deep learning based solution techniques has become attractive in recent years due to their inherent capability in handling data and their advantage as surrogate models.

The deep learning based PDE solvers rely on the Fourier NN [5] and universal approximation theorem for neural networks [6]. A standard multilayer perceptron approximating a given function with arbitrary error is one of the earliest efforts to solve Ordinary Differential Equations (ODEs) and PDEs using a deep neural network (DNN) [7]. Recently, Berg and Nyström [8] extended this idea to domains with complex geometries.

Han et al. [9] reformulated PDEs as backward stochastic PDEs and approximated the gradient of the solution using DNNs. They were able to solve high dimensional PDEs and overcome the curse of dimensionality. A new paradigm shift for solving PDEs using DNNs emerged with the development of Physics Informed Neural Network (PINN) [10]. PINN benefits from the performance and accuracy of automatic differentiation[11]. The inputs of the NN are space-time independent variables and the partial derivatives of the output with respect to inputs are compiled into a physics-based loss term in the form of a PDE.

† These authors contributed equally to this work.
[1]Corresponding author. Tel.: +1 520 621 6113.
*E-mail addresses:* amavi@arizona.edu (A. Mavi), acbekar@arizona.edu (A. C. Bekar), ehsanh@mit.edu (E. Haghighat), madenci@arizona.edu (E. Madenci),

If available, known values of the field are included in the loss term, and initial and boundary conditions are enforced by adding corresponding terms to the loss in the form of mean squared error. The parameters of the feed forward NN are optimized by minimizing these losses. PINNs show extraordinary performance for interpolation and surrogate modeling. They have been successfully applied to solve many problems [12−18]. However, the training of PINNs still presents challenges despite the extensive improvements by Wang et al. [19−21] and Chen et al. [22]. Also, the convergence of PINNs can be slow for complex problems with multiple governing equations. Another issue is the extrapolation capability of PINNs. Even though PINNs are successful interpolators, their extrapolation performance is rather poor. Predicting a solution outside the training time interval gives unsatisfactory results [23]. A recent work from Lu et al. [24] introduces DeepOnets which is based on learning the underlying operators of the governing equations. Oommen et al. [25] proposed a hybrid autoencoder network to predict the time evolution of two-phase microstructures by learning the dynamics in latent space using a DeepOnet architecture.

Recently, significant attention is concentrated on recurrent learning based PDE solvers because of their robustness for time-extrapolation [23, 26, 27]. One of the fundamental recurrent learning structures used for solving PDEs is Recurrent Neural Networks (RNNs). They are designed to capture long-time behavior of time-series inputs [28]. However, their training is challenging because of vanishing or exploding gradients[29]. Long-short Term Memory (LSTM) structure is proven to be easier to train compared to RNNs due to their ingenious gated structure [30]. However, the basic LSTM architecture is a fully-connected architecture and it does not conserve spatio-temporal dependencies. On the other hand, NNs with convolutional layers have proven to perform better on data sets with spatial dependencies and easier to train [31]. Recently, Shi et al. [32] introduced Convolutional LSTM (ConvLSTM) architecture and demonstrated its superior performance to LSTMs for capturing data with spatio-temporal correlations. Ren et al. [23] proposed a PDE solver using ConvLSTM by employing FDM to estimate physical losses.

Extending the work of Ren et al. [23], we present a novel unsupervised convolutional NN architecture with nonlocal interactions for solving PDEs. Instead of FDM, we enforce the physical governing equations using Peridynamic Differential Operator (PDDO) [33] as convolutional filters. Unlike FDM, PDDO accounts for nonlocal phenomena with arbitrary horizon size. Therefore, the filters can be generic and of any size.

Our neural network has encoder-decoder layers with a ConvLSTM layer between them for capturing time-dynamics in a smaller latent space. Inspired by reduced order modeling [34, 35], we also enforce the physical losses as a constraint in the latent (reduced) space. Furthermore, we modify the ConvLSTM architecture by employing a novel activation function [36] to improve the predictive capability of the present learning architecture for physics with periodic behavior. Finally, we demonstrate its improved training performance and better extrapolation capability by comparing against PINN type solvers.

This study is organized as follows. In Section 2, we explain the specific form of the PDEs and provide the details of the present NN architecture and the solution strategy. In Section 3, we briefly describe the PDDO and explain the structure of the derivative filters. In Section 4, we present numerical experiments and report performance comparison between PINN and the present NN architecture. Finally, we discuss the results and summarize the main conclusions in Section 5.

## 2. Problem Statement and Methodology

In this study, we consider the PDE systems in the following form

$$\mathbf{u}_t + \mathcal{N}[\mathbf{u}] = 0, \quad \mathbf{x} \in \Omega, \quad t > 0 \tag{2.1a}$$

$$\mathbf{u}(\mathbf{x}, t) \in \mathbb{R}_{per} \quad \forall t > 0 \tag{2.1b}$$

and

$$\mathbf{u}(\mathbf{x}, 0) = \mathbf{u}_0(\mathbf{x}) \tag{2.1c}$$

where $\circ_t$ denotes partial derivative with respect to time and $\mathcal{N}[\circ]$ is a nonlinear differential operator acting on $\mathbf{u}$. The physical domain is denoted by $\Omega$ and $\mathbb{R}_{per}$ indicates that the solution to the PDE satisfies periodicity in $\mathbb{R}^d$, i.e. the solution satisfies periodic boundary conditions and $\mathbf{u}_0(\mathbf{x})$ is the initial condition. In a 2-dimensional domain, the unknown vector includes 2 dependent variables, $\mathbf{u} = \{u, v\}$. The shape of the domain is square with a uniform discretization.

### 2.1 Network Architecture

Motivated by the work of Ren et al. [23], our neural network depicted in Figure 1 consists of 3 main components: 1) convolutional encoder, 2) ConvLSTM, and 3) convolutional decoder. The encoder part projects a given input to a smaller latent space by extracting the hidden representations. The ConvLSTM part encodes the time dynamics of the process between time steps. The decoder part expands the output of the ConvLSTM to the same dimensions as the input of the NN. Subsequently, this output is multiplied with the time step of $\delta t$ and added to the input which presents the integrated field value at the next time step, i.e., $\mathbf{u}_i = \mathbf{u}_{i-1} + \delta t \mathcal{N}[\mathbf{u}_{i-1}, \lambda]$.

Therefore, this neural network functions as a time integrator. Using the discretely integrated field variables, we construct the time step loss by imposing the discretized PDEs. Requiring that physical governing equations are satisfied at any scale, the present architecture also constrains the latent space to satisfy the governing relations. Therefore, the learning process is unsupervised. This process is further explained in Section 3.

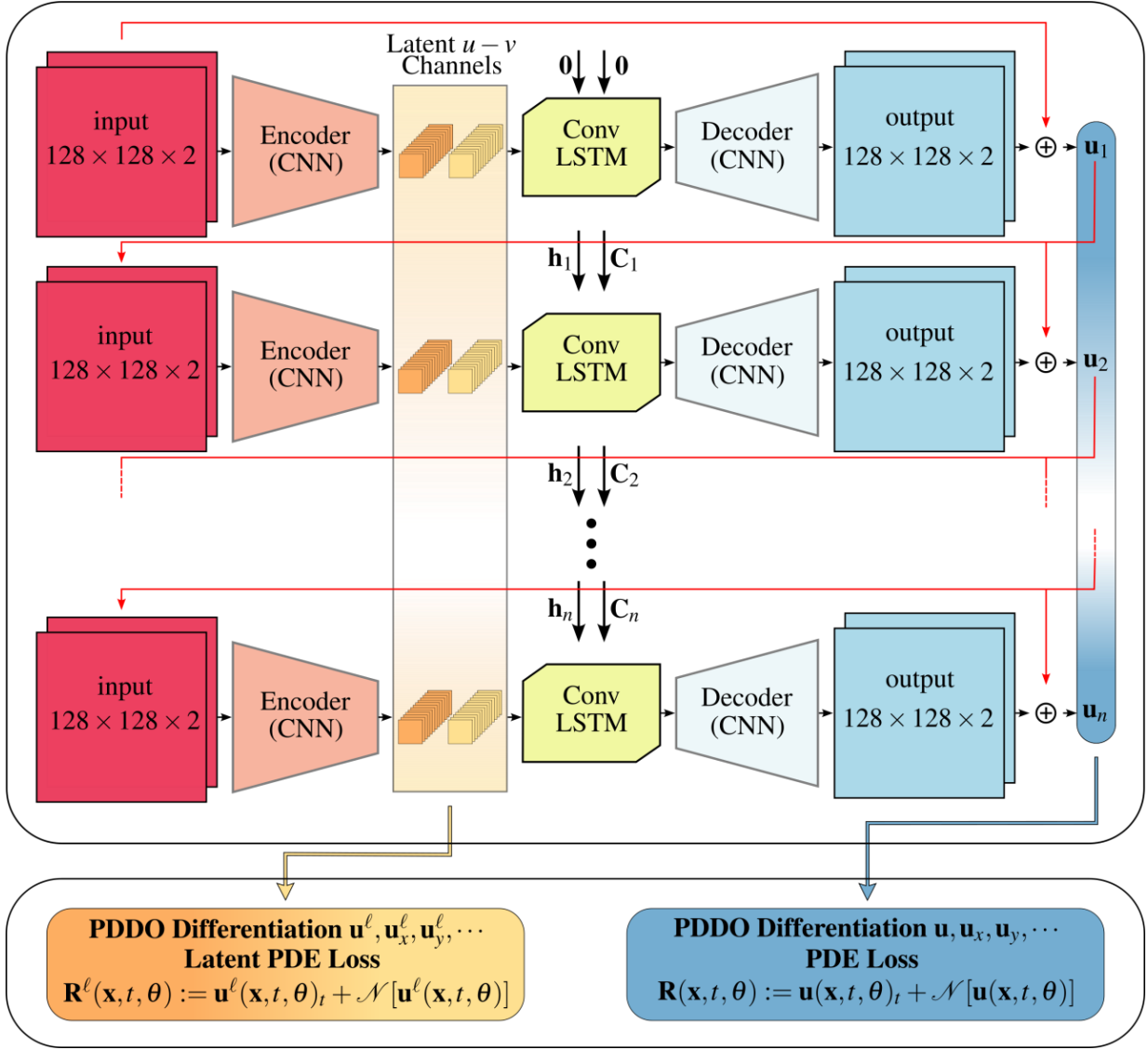

**Figure 1.** Proposed network architecture

## 2.2 The encoder-decoder architecture

As shown in Figure 2, the encoder consists of 3 convolutional layers with hyperbolic-tangent activation functions, and 1×1 periodic padding. First layer uses 8 filters with 4×4 kernels and 2 strides, second layer has 32 filters with 4×4 kernels and 2 strides and the third layer has 64 filters with 4×4 kernels and 2 strides. The output tensor of these three layers has dimensions of 16×16×64. This tensor is split into two tensors with dimensions of 16×16×32. These two tensors contain the lower dimensional representations of field variables. The physics informed PDE residual is enforced separately as explained in Section 2.4. Unlike the architecture proposed by Ren et al. [23], the present structure compresses the input using the autoencoder, i.e., extracts smaller features in the latent space. This modification increases the generalization capability of the NN by preventing the NN from memorizing the field data, i.e., overfitting. This modification also shrinks the input data so that the computational burden on ConvLSTM decreases. Figure 2 shows the structure of the encoder part of the present NN. It has 3 layers with hyperbolic-tangent activation functions and 1×1 periodic padding. First layer uses 64 filters with 3×3 kernels and 1 stride, second layer has 32 filters with 3×3 kernels and 1 stride and the third layer has 8 filters with 3×3 kernels and 1 stride. The output tensor has dimensions of 128×128×2 so that the input size is recovered. Figure 3 shows the structure of the encoder part of the NN. Kernel is the applied filter to a channel with constant size and it extracts the features of a channel. Padding is the extent of pixels added to the channel edges when the kernel is applied. Stride is the level of shift (movement) of a filter while calculating the convolution between the channel and the kernel.

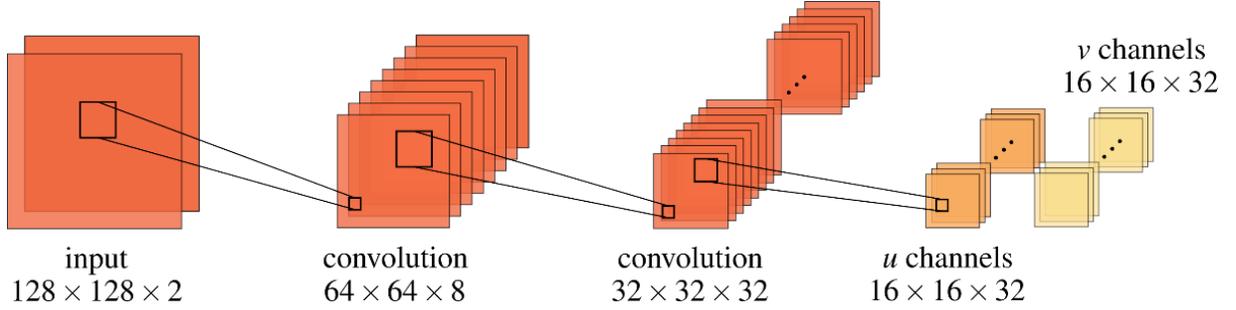

**Figure 2.** Encoder architecture and tensor sizes

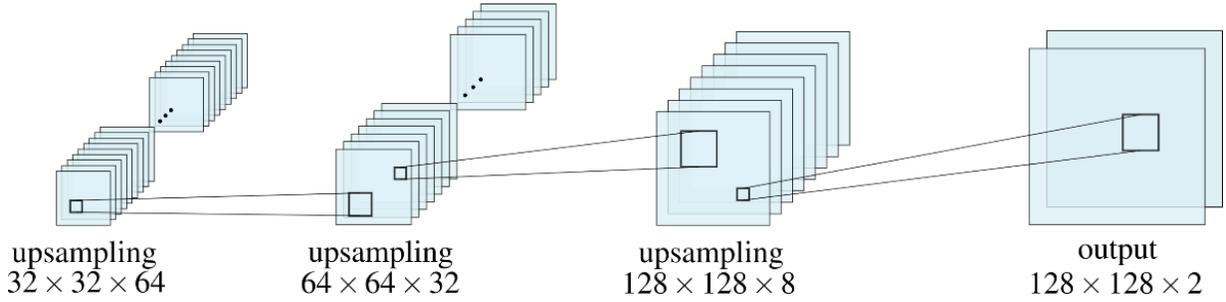

**Figure 3.** Decoder architecture and tensor sizes

## 2.3 Periodic Convolutional LSTM

As depicted in Figure 4, the convolutional LSTM cell architecture is expressed mathematically as follows:

$$\mathbf{i}_t = \sigma(W_{xi} * \mathbf{x}_t + W_{hi} * \mathbf{h}_{t-1} + W_{ci} \circ \mathbf{C}_{t-1} + b_i) \tag{2.2a}$$

$$\mathbf{f}_t = \sigma(W_{xf} * \mathbf{x}_t + W_{hf} * \mathbf{h}_{t-1} + W_{cf} \circ \mathbf{C}_{t-1} + b_f) \tag{2.2b}$$

$$\tilde{\mathbf{C}}_t = \Xi_\alpha(W_{xc} * \mathbf{x}_t + W_{hc} * \mathbf{h}_{t-1} + b_c) \tag{2.2c}$$

$$\mathbf{C}_t = \mathbf{f}_t \circ \mathbf{C}_{t-1} + \mathbf{i}_t \circ \tilde{\mathbf{C}}_t \tag{2.2d}$$

$$\mathbf{o}_t = \sigma(W_{xo} * \mathbf{x}_t + W_{ho} * \mathbf{h}_{t-1}) \qquad (2.2e)$$

and

$$\mathbf{h}_t = \mathbf{o}_t \circ \tanh(\mathbf{C}_t) \qquad (2.2f)$$

in which, $\mathbf{x}_t$ is the input, $\mathbf{C}_t$ is the cell state, $\tilde{\mathbf{C}}_t$ is the internal cell state, $\mathbf{h}_t$ is the hidden state, $\mathbf{i}_t$ is the input gate, $\mathbf{f}_t$ is the forget gate and $\mathbf{o}_t$ is the output gate. The operations, '$*$' and '$\circ$' denote the convolution operation and element-wise multiplication (Hadamard product), respectively.

Forget gate decides whether to pass the information to the cell state or to "forget" it by filtering that information. Input gate decides on the information added to the cell state from the internal cell state. Output gate decides on the features passed to the hidden state from the updated cell state.

To account for periodicity, we employ the periodic activation function $\Xi_\alpha(x) = x + \frac{1}{\alpha}\sin^2(\alpha x)$ [36], instead of the commonly accepted hyperbolic-tangent function. It is worth noting that $\alpha$ is treated as a trainable parameter rather than with a fixed value as originally introduced [36]. It is initialized randomly between $-2\pi$ and $2\pi$.

This modification helps the network learn faster when periodic solutions are present. It is also capable of representing nonperiodic and decaying behavior. Appendix B contains further information about the periodic activation function. ConvLSTM has kernel sizes of 3×3 with 1×1 periodic padding. Figure 4 depicts the structure of the novel ConvLSTM cell.

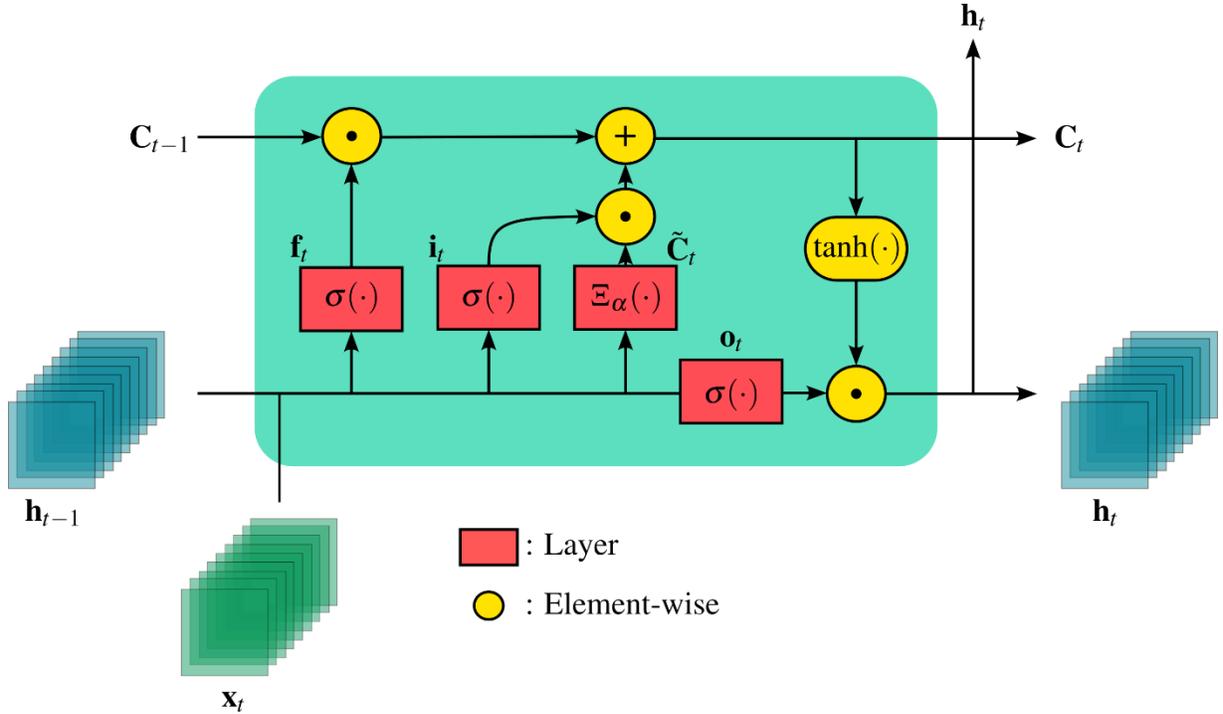

**Figure 4.** The modified ConvLSTM cell

## 2.4 Output and Latent PDE Losses

The learning algorithm is unsupervised, i.e., we do not use any labeled data. However, we use the governing equation of the field to construct the loss. The output (standard) loss has the following form

$$\mathbf{R}(x,t;\boldsymbol{\theta}) := \mathbf{u}(\mathbf{x},t;\boldsymbol{\theta})_t + \mathcal{N}[\mathbf{u}(\mathbf{x},t;\boldsymbol{\theta})] \tag{2.3}$$

where $\boldsymbol{\theta}$ represents the NN parameters and $\mathcal{N}[\circ]$ represents the differential operator which often include derivatives and products of derivatives.

Motivated by the fact that physical laws hold true at any scale, we also constrain the latent (reduced) space with a similar PDE-loss as shown in Figure 1. The loss is similar to Eq. (2.3); however, it acts on a different scale and only affects the encoding part of the NN. The latent loss has the following form

$$\mathbf{R}^\ell(x,t;\boldsymbol{\theta}) := \mathbf{u}^\ell(\mathbf{x},t;\boldsymbol{\theta})_t + \mathcal{N}[\mathbf{u}^\ell(\mathbf{x},t;\boldsymbol{\theta})] \tag{2.4}$$

where $\circ^\ell$ denotes the quantity in latent space. The remaining challenge is to evaluate a loss function that includes differential operators. Among NN based solvers, PINN uses automatic differentiation to calculate the partial derivatives of the field variable of desired order [10]. PhyCRNet uses convolutional filters based on FDM to evaluate the partial derivatives [17−23]. As discussed in detail in the next section, we use the Peridynamic Differential Operator (PDDO) [33] to construct the derivative filters. Unlike FDM that can only account for interaction between points in close proximity, PDDO can handle an arbitrary filter size. Hence, it is suitable for both local and nonlocal approximations. Therefore, it can be viewed as a generalization of PhyCRNet.

## 3. Nonlocal Convolutional Derivative Filters using PDDO

To evaluate derivatives at a point of interest $\mathbf{x}$, PDDO uses a family of points, $\mathbf{H_x}$ shown in Figure 5. It is defined as $\mathbf{H_x} = \{\mathbf{x}' | w(\mathbf{x}' - \mathbf{x}) > 0\}$. Each point $\mathbf{x}$ has its own unique family in its domain of interaction. The non-dimensional weight function defining the degree of interaction between points is defined as

$$w(|\boldsymbol{\xi}|) = e^{-4|\boldsymbol{\xi}|^2/\delta^2} \tag{3.1}$$

where $\boldsymbol{\xi} = \mathbf{x}' - \mathbf{x}$ is the relative position of a point $\mathbf{x}'$ with the point of interest, $\mathbf{x}$. The parameter $\delta$, a.k.a, the horizon, defines the extent of the interaction domain (long-range interactions). In discrete form, the family members of point $\mathbf{x}$ are denoted as $\mathbf{H_x} = \left(\mathbf{x}_{(1)}, \mathbf{x}_{(2)}, \ldots, \mathbf{x}_{(N)}\right)$, and their relative positions are defined as $\boldsymbol{\xi}_{(j)} = \mathbf{x}_{(j)} - \mathbf{x}$.

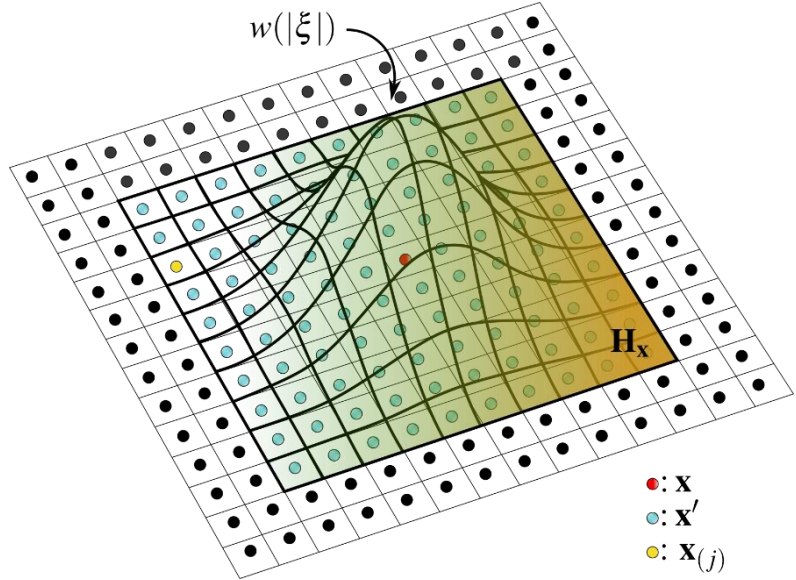

**Figure 5.** Interaction domain, $\mathbf{H_x}$ of point $\mathbf{x}$ includes its family members, $\mathbf{x}_{(j)}$

As detailed by Madenci et al. [33], the derivation of PDDO employs Taylor Series Expansion (TSE), weighted integration and orthogonality conditions (see Appendix A). Hence, the nonlocal PD representation of function $f(\mathbf{x})$ and its first and second order derivatives in 2D, using 2nd order Taylor Series Expansion (TSE), can be expressed in continuous and discrete forms as

$$f(\mathbf{x}) = \int_{\mathbf{H_x}} f(\mathbf{x}+\boldsymbol{\xi}) g_2^{00}(\boldsymbol{\xi}) dA \approx \sum_{\mathbf{x}_{(j)} \in \mathbf{H_x}} f(\mathbf{x}_{(j)}) g_2^{00}(\boldsymbol{\xi}_{(j)}) A_{(j)}, \qquad (3.2)$$

$$\begin{Bmatrix} f_x \\ f_y \end{Bmatrix} = \int_{\mathbf{H_x}} f(\mathbf{x}+\boldsymbol{\xi}) \begin{Bmatrix} g_2^{10}(\boldsymbol{\xi}) \\ g_2^{01}(\boldsymbol{\xi}) \end{Bmatrix} dA \approx \sum_{\mathbf{x}_{(j)} \in \mathbf{H_x}} f(\mathbf{x}_{(j)}) \begin{Bmatrix} g^{10}(\boldsymbol{\xi}_{(j)}) \\ g^{01}(\boldsymbol{\xi}_{(j)}) \end{Bmatrix} A_{(j)}, \qquad (3.3)$$

and

$$\begin{Bmatrix} f_{xx} \\ f_{yy} \\ f_{xy} \end{Bmatrix} = \int_{\mathbf{H_x}} f(\mathbf{x}+\boldsymbol{\xi}) \begin{Bmatrix} g_2^{20}(\boldsymbol{\xi}) \\ g_2^{02}(\boldsymbol{\xi}) \\ g_2^{11}(\boldsymbol{\xi}) \end{Bmatrix} dA \approx \sum_{\mathbf{x}_{(j)} \in \mathbf{H_x}} f(\mathbf{x}_{(j)}) \begin{Bmatrix} g^{20}(\boldsymbol{\xi}_{(j)}) \\ g^{02}(\boldsymbol{\xi}_{(j)}) \\ g^{11}(\boldsymbol{\xi}_{(j)}) \end{Bmatrix} A_{(j)}, \qquad (3.4)$$

where $g_2^{p_1 p_2}(\boldsymbol{\xi})$ with $(p,q = 0,1,2)$ represent the PD functions obtained by enforcing the orthogonality condition of PDDO [33−37], and the integration is performed over the interaction domain.

## 3.1 PDDO-CNN Filters

Discrete summation in Eqs. (3.2-3.4) can be considered as a convolution operation. The PD functions, $g_2^{p_1 p_2}(\xi)$ form the basis of the convolutional filters. Figure 6 depicts the PD integration as a convolutional operation.

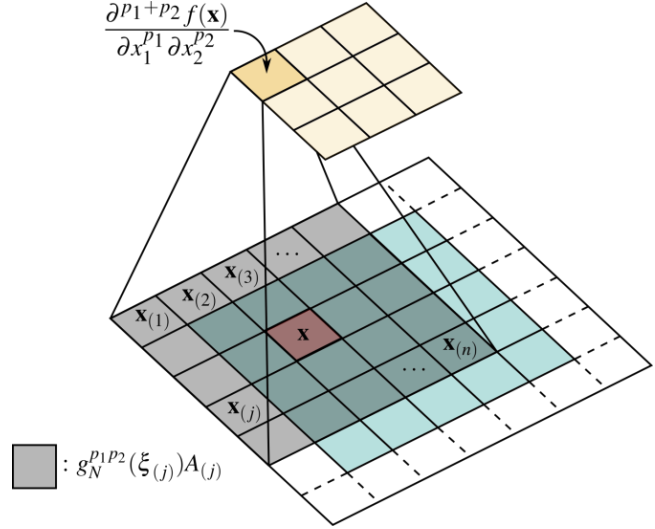

**Figure 6.** The general form of Peridynamic convolutional filters

The PDDO filters are intrinsically more efficient because of the arbitrary size of spatial area they cover. Figure 7 shows the difference between PDDO filters and FDM filters. Because of the locality of the FDM filters, most of the values in the convolutional filter are zero. On the other hand, due to nonlocality, PDDO filters have mostly non-zero values; thus, they incorporate more information from the surrounding pixels or family member values. Hence, PDDO derivative filters are genuine convolutional filters.

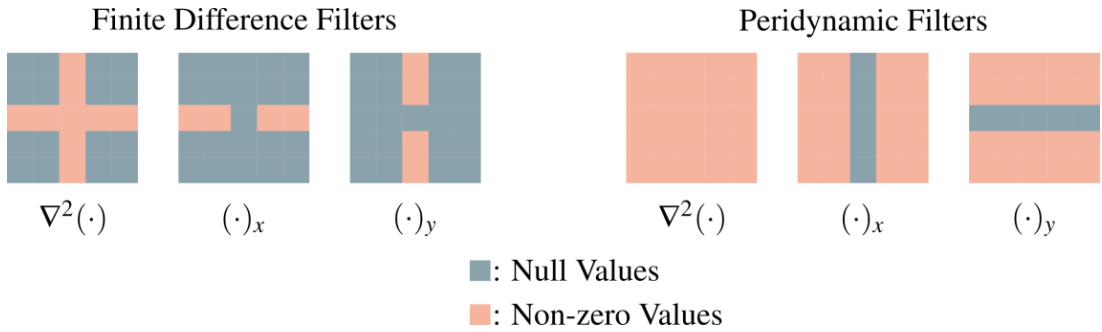

**Figure 7.** Comparison of sparsity for PDDO and FDM filters

The spatial derivatives are evaluated using PDDO-Convolutional filters. To determine the time derivative at the output, we use a filter based on a second order finite difference scheme as

$$\circ_t \approx [-1, 0, 1] \times \frac{1}{2\Delta t} \tag{3.5}$$

Consequently, all necessary spatial and temporal filters are established to evaluate the derivatives in the output and latent loss functions.

## 4. Numerical results

This section consists of the numerical experiments conducted to test the present neural network architecture. It presents solutions to 2D Coupled Burgers' equation, $\lambda-\omega$ Reaction-Diffusion equation and Gray-Scott equation. Training losses for these numerical experiments are shown in Appendix C. We use the same discretization for all cases and compare our interpolation and extrapolation results with those of PINN. The solutions are constructed on (AMD EPYC 7642 48-core) CPU with 2.4 GHz clock speed and 64 GB RAM. The code for our network architecture is written using TensorFlow library, and it is available at: https://github.com/ardamavi/PI-rCNN.

### 4.1 Burgers' Equation

Two-dimensional coupled nonlinear Burgers' equation is commonly used to verify numerical solvers. It can model traffic flow, hydrodynamic turbulence, vorticity transport, fluid flow with shock and numerous other phenomena. Burgers' equation in vector form can be expressed as

$$\mathbf{u}_t + \mathbf{u} \cdot \nabla \mathbf{u} - \nu \Delta \mathbf{u} = 0 \tag{4.1}$$

where $\nabla(\circ) = \{(\circ)_x, (\circ)_y\}$, $\Delta(\circ) = (\circ)_{xx} + (\circ)_{yy}$, $\mathbf{u} = \{u, v\}$ is the vector of fluid velocity and $\nu$ is the kinematic viscosity. It can be rewritten as

$$u_t = -uu_x - vu_y + \nu \Delta u \tag{4.2a}$$

and

$$v_t = -uv_x - vv_y + \nu \Delta v \tag{4.2b}$$

in which subscript denotes differentiation. We chose the spatial domain for this problem as $\mathbf{x} \in [0,1] \times [0,1]$. Initial condition is randomly generated using $\mathbf{u}_0 \sim \mathcal{N}(0, 625(-\Delta + 25\mathbf{I})^{-2})$ [23]. Our network is trained for $t \in [0,2]s$ using 1000 time steps with learning rate of $0.001$; it is updated by using adaptive learning rate. Training time is 28.51 hours. Figure 8 depicts the training outputs and absolute errors for present architecture and PINN. The absolute error for training domain is relatively small for PINN because of the hard constraint of initial conditions; however, the error for PINN gradually increases as time increases. It is worth noting that a decay in accuracy does not occur in our method due to the specification of initial condition as input. Figure 9 shows the results for extrapolation predictions from present architecture and PINN for $t \in [2,3]s$. Pairs of columns in Figs. 8 and 9 show $u$ and $v$ fields, respectively. The present model performs better with more accurate extrapolation results compared to PINN. Also, the error in our method decreases as time increases while PINN error consistently increases.

The reason for this superior extrapolation performance compared to PINN is its ability to generalize the dynamics of the field. This capability is due to its architecture. It projects the input to a smaller latent space using an encoder structure and learns the time evolution in the smaller latent space. This learning involves two key features: first one is the convolutional LSTM, which excels at forecasting time dynamics, and the second one is the latent physics loss.

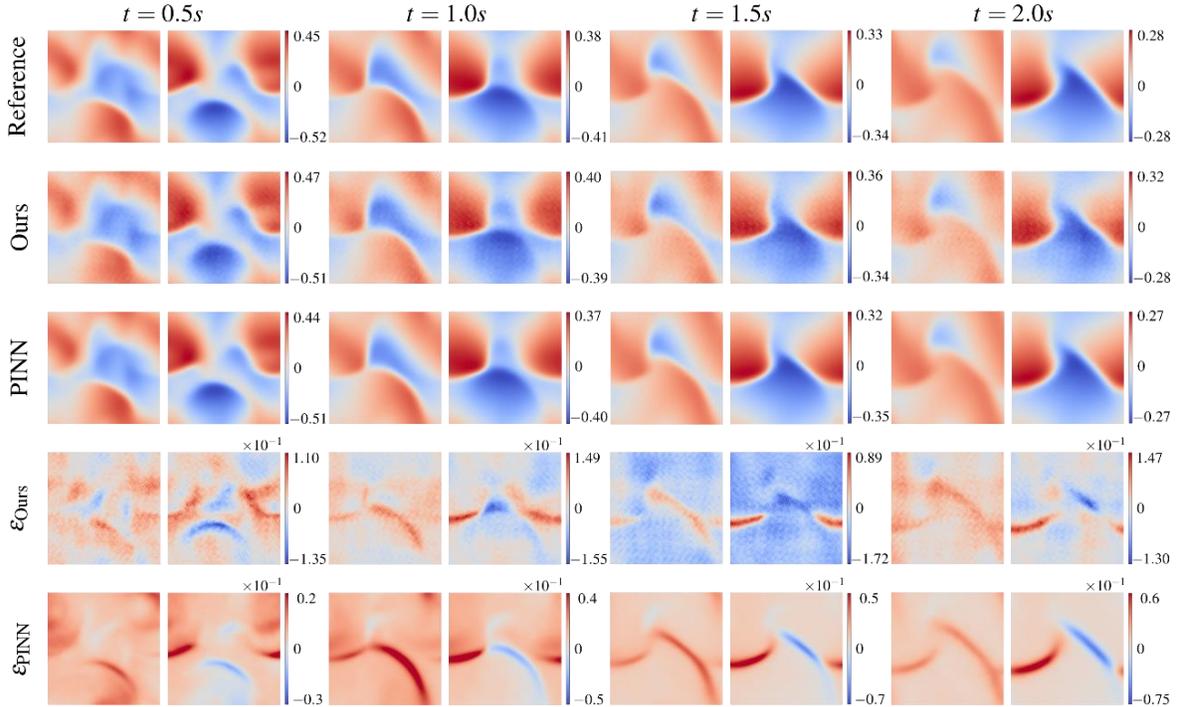

**Figure 8.** Training results for Burgers' equation at $t = [0.5, 1.0, 1.5, 2.0]s$

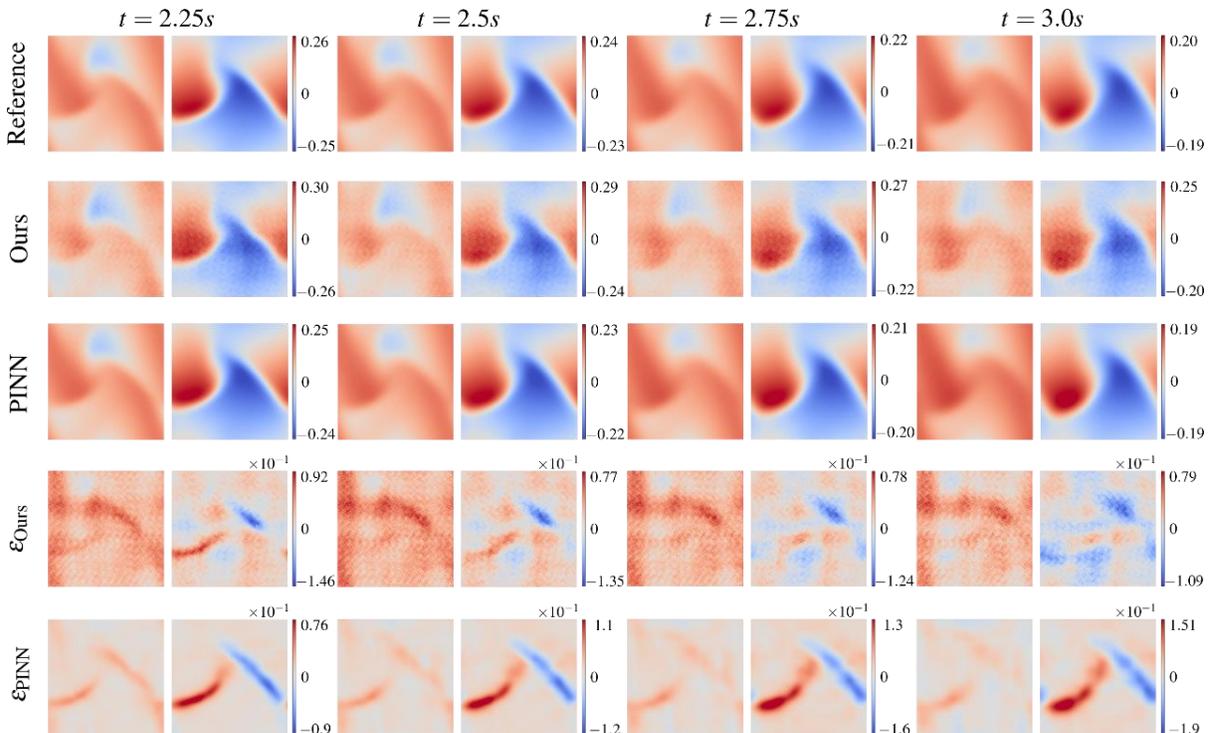

**Figure 9** Testing results for Burgers' equation at $t = [2.25, 2.5, 2.75, 3.0]s$

## 4.2 $\lambda-\omega$ Reaction-Diffusion Equation

$\lambda-\omega$ reaction diffusion equation is a coupled oscillatory differential equation. It is a special case of Ginzburg-Landau equation [38–39]. It generates stable travelling waves and has a limit cycle for certain $\lambda$ and $\omega$ values. A 2D $\lambda-\omega$ reaction diffusion equation can be written as

$$u_t = 0.1\Delta u + \lambda(r)u - \omega(r)v \tag{4.3a}$$

and

$$v_t = 0.1\Delta v + \omega(r)u + \lambda(r)v \tag{4.3b}$$

Where $\Delta(\circ) = (\circ)_{xx} + (\circ)_{yy}$, $r = u^2 + v^2$, $\lambda(r) = 1 - r^2$ and $\omega(r) = -r^2$. For the solution, we chose $\mathbf{x} \in [-10,10] \times [-10,10]$ and employ the dataset provided by Rudy et al. [40]. Their solution is simulated using spectral method. We train our model for $t \in [0,5]s$ using 400 time steps. Learning rate at the beginning is specified as 0.001 and updated using adaptive learning rate. Training time is specified as 15.6 hours. Figure 10 shows the training results of the current model and PINN for $\lambda-\omega$ reaction diffusion equation. Our architecture performs again better in interpolation compared to PINN and has a smaller absolute error for $\lambda-\omega$ reaction diffusion equation. Figure 11 shows the predictions from the present approach and PINN for $t \in [5,10]s$. Pairs of columns in Figs. 10 and 11 show $u$ and $v$ fields, respectively. Our model performs better in extrapolation with a smaller absolute error than that of PINN.

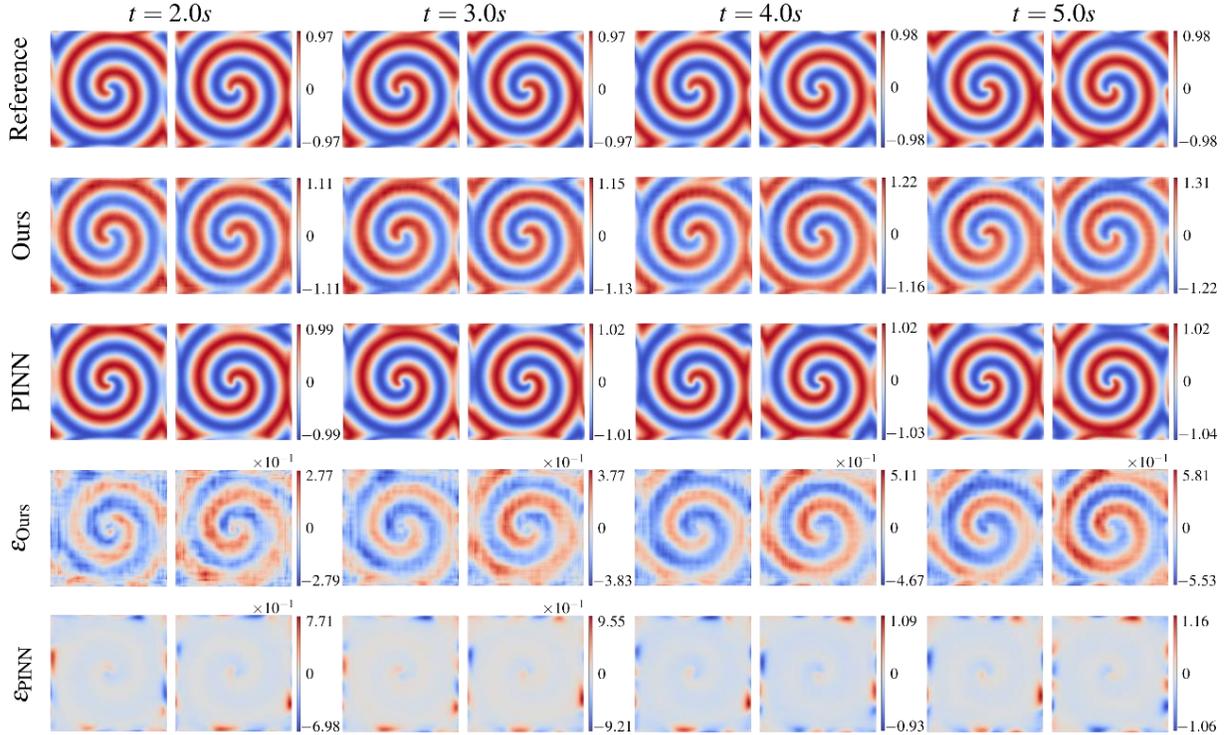

**Figure 10** Training results, $\lambda-\omega$ reaction diffusion equation $t=[2.0, 3.0, 4.0, 5.0]s$

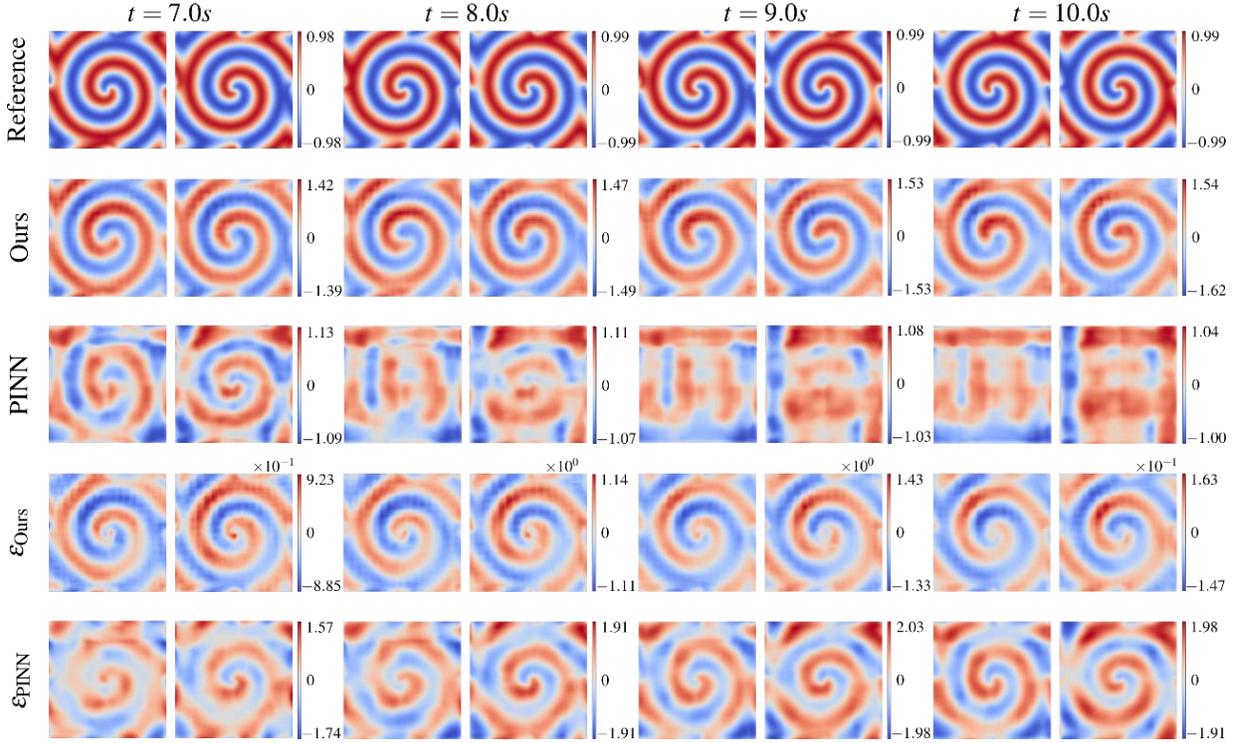

**Figure 11** Testing results, $\lambda-\omega$ reaction diffusion equation $t=[7.0, 8.0, 9.0, 10.0]s$

### 4.3 Gray-Scott Equation

Gray-Scott equation is a coupled reaction-diffusion type equation which models the following chemical reactions

$$\mathcal{U} + 2\mathcal{V} \to 3\mathcal{V} \tag{4.4a}$$

and

$$\mathcal{V} \to \mathcal{P} \tag{4.4b}$$

where $\mathcal{V}$ is a catalyst, $\mathcal{P}$ is an inert product and $\mathcal{U}$ is a chemical species. Gray-Scott equation can be expressed as

$$u_t = \epsilon_1 \Delta u + b(1-u) - uv^2 \tag{4.5a}$$

and

$$v_t = \epsilon_2 \Delta v - dv + uv^2 \tag{4.5b}$$

Where $\Delta(\circ) = (\circ)_{xx} + (\circ)_{yy}$, $u(x,y,t)$ and $v(x,y,t)$ denote the concentrations for $\mathcal{U}$ and $\mathcal{V}$, $b$ is the rate of $\mathcal{U}$ feed, $\epsilon_1$ and $\epsilon_2$ are diffusivity coefficients and $d$ is the rate of $\mathcal{V}$ transforming to $\mathcal{P}$. For this specific case, we chose $\mathbf{x} \in [-0.2, 0.2] \times [-0.2, 0.2]$, $\epsilon_1 = 0.00002$, $\epsilon_2 = 0.00001$, $b = 0.04$, $d = 0.1$ and use **chebfun** package to obtain simulation results for comparison [41]. We train our network for $t \in [0,10]s$ using 1000 time steps. Learning rate at the beginning is 0.001 and updated using adaptive learning rate. Training time is specified as 23.84 hours, and solution is predicted for $t \in [10,15]s$. Figure 12 shows the training and testing results of current method for Gray-Scott equation. Figure 13 compares the extrapolation predictions of the present model with those of PINN for $t \in [10,15]s$. Pairs of columns in Figs. 12 and 13 show $u$ and $v$ fields, respectively. The present model predictions have a smaller absolute error than those of PINN. However, the training errors are smaller for PINN than those of current method because of hard imposition of initial condition.

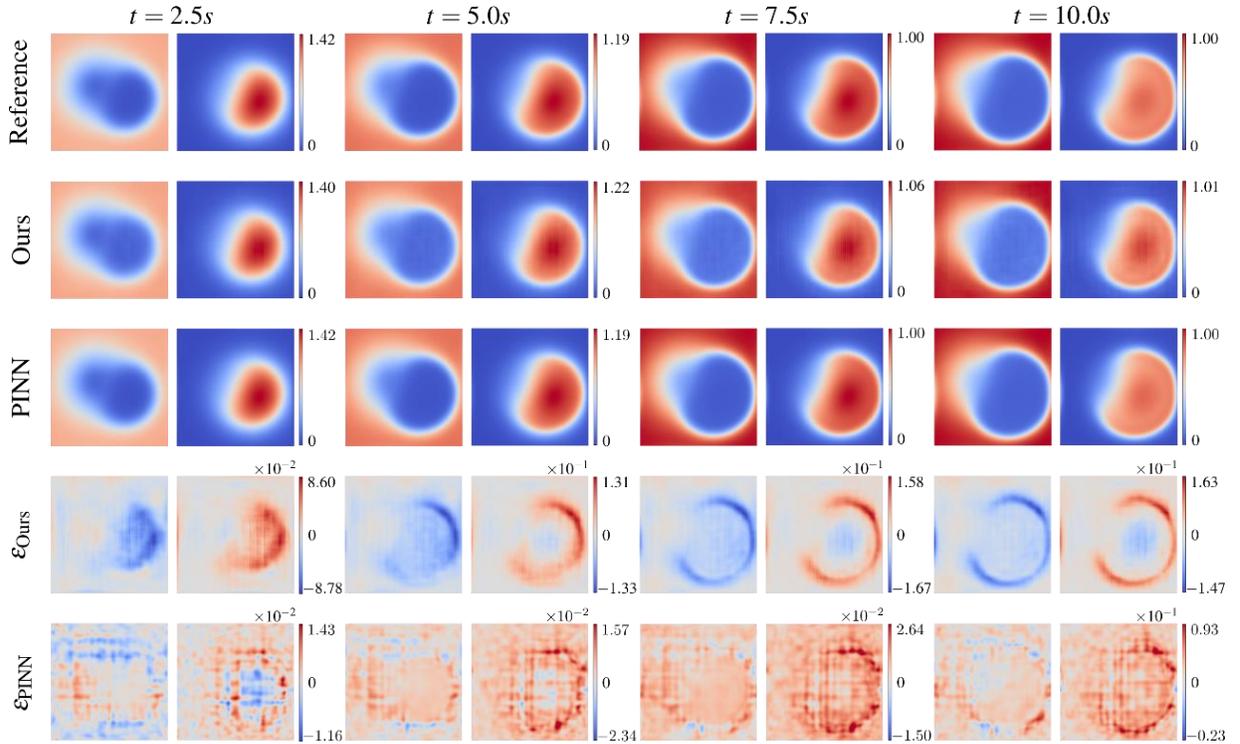

**Figure 12** Training results, Gray-Scott equation for $t = [2.5, 5.0, 7.5, 10.0]s$

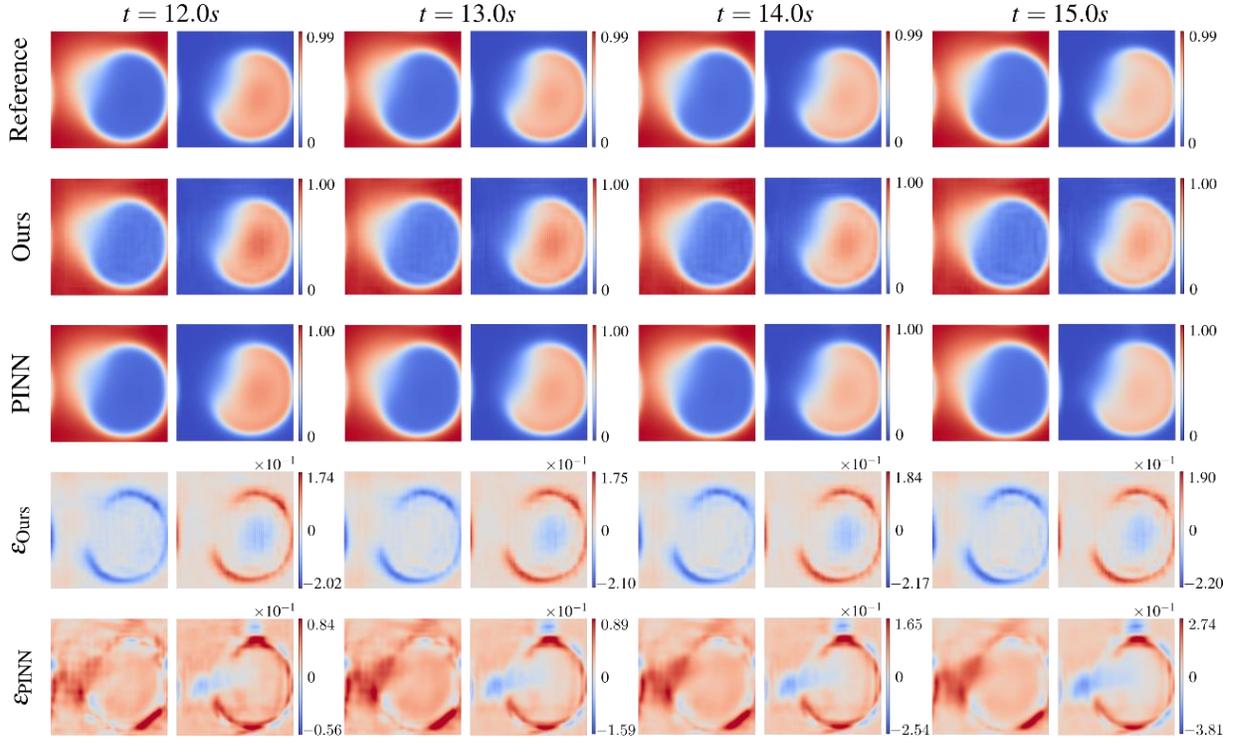

**Figure 13** Testing results, Gray-Scott equation for $t = [12.0, 13.0, 14.0, 15.0]\,s$

## 5. Conclusions

This study presents a novel deep learning architecture to obtain the forward solution for coupled PDEs by employing only convolutional layers. The main reason for this choice is that the convolutional layers have sparse connections with more robustness to overfitting. The solutions to PDEs contain spatial dependencies; therefore, convolutional layers are more appropriate for capturing such dependencies and provide better generalization performance.

The classical methods for solving PDEs do not offer any generalization property. We use a ConvLSTM to capture the time dynamics and enhance the extrapolation capability of the neural network. Its training is unsupervised with physics informed in the form of governing equation at the output. Also, we impose the conservation relations on the latent space; therefore, we reach higher accuracy compared to other similar architectures.

We use PDDO to obtain the nonlocal convolutional filters for derivative calculation of the field variable. This choice improves the efficiency of the derivative filters and enables us to choose the filter size freely. We compare our results with PINN approach and observe that our method is more capable of extrapolating the solution for future time steps than PINN. One limitation of the present deep learning architecture is the generalization of its performance for different initial conditions. This drawback can be handled by using a generative model at encoding-decoding layers.

**Acknowledgement**
EM, ACB and AM performed this work as part of the ongoing research at the MURI Center for Material Failure Prediction through Peridynamics at the University of Arizona (AFOSR Grant No. FA9550-14-1-0073).

**Appendix A – Peridynamic differential operator**

In a 2-dimensional space, a function, $f(\mathbf{x}+\boldsymbol{\xi})$ can be expressed in terms of Taylor Series Expansion (TSE) as

$$f(\mathbf{x}+\boldsymbol{\xi}) = f(\mathbf{x}) + \xi_1 \frac{\partial f(\mathbf{x})}{\partial x_1} + \xi_2 \frac{\partial f(\mathbf{x})}{\partial x_2} + \frac{1}{2!}\xi_1^2 \frac{\partial^2 f(\mathbf{x})}{\partial x_1^2} + \frac{1}{2!}\xi_2^2 \frac{\partial^2 f(\mathbf{x})}{\partial x_2^2} + \xi_1\xi_2 \frac{\partial^2 f(\mathbf{x})}{\partial x_1 \partial x_2} + \mathcal{R} \quad (A.1)$$

where $\mathcal{R}$ is the remainder. Multiplying each term with PD functions, $g_2^{p_1 p_2}(\boldsymbol{\xi})$ and integrating over the domain of interaction (family), $\mathbf{H}_\mathbf{x}$ results in

$$\int_{\mathbf{H}_\mathbf{x}} f(\mathbf{x}+\boldsymbol{\xi}) \, g_2^{p_1 p_2}(\boldsymbol{\xi}) dV = \int_{\mathbf{H}_\mathbf{x}} f(\mathbf{x}) \, g_2^{p_1 p_2}(\boldsymbol{\xi}) dV + \frac{\partial f(\mathbf{x})}{\partial x_1} \int_{\mathbf{H}_\mathbf{x}} \xi_1 \, g_2^{p_1 p_2}(\boldsymbol{\xi}) dV$$

$$+ \frac{\partial f(\mathbf{x})}{\partial x_2} \int_{\mathbf{H}_\mathbf{x}} \xi_2 \, g_2^{p_1 p_2}(\boldsymbol{\xi}) dV + \frac{1}{2}\frac{\partial f^2(\mathbf{x})}{\partial x_1^2} \int_{\mathbf{H}_\mathbf{x}} \xi_1^2 \, g_2^{p_1 p_2}(\boldsymbol{\xi}) dV$$

$$+ \frac{1}{2}\frac{\partial f^2(\mathbf{x})}{\partial x_2^2} \int_{\mathbf{H}_\mathbf{x}} \xi_2^2 \, g_2^{p_1 p_2}(\boldsymbol{\xi}) dV + \frac{\partial^2 f(\mathbf{x})}{\partial x_1 \partial x_2} \int_{\mathbf{H}_\mathbf{x}} \xi_1 \xi_2 \, g_2^{p_1 p_2}(\boldsymbol{\xi}) dV$$

(A.2)

in which the point $\mathbf{x}$ is not necessarily symmetrically located in the domain of interaction. The initial relative position, $\boldsymbol{\xi}$, between points $\mathbf{x}$ and $\mathbf{x}'$ can be expressed as $\boldsymbol{\xi} = \mathbf{x}' - \mathbf{x}$. This ability permits each point to have its own unique family with an arbitrary position. Therefore, the size and shape of each family can be different, and they significantly influence the degree of nonlocality. In general, the family of a point can be nonsymmetric due to nonuniform spatial discretization.

The degree of interaction between the material points in each family is specified by a nondimensional weight function, $w(|\boldsymbol{\xi}|)$, which can vary from point to point. The interactions become more local with decreasing family size. Thus, the family size and shape are important parameters. Each point occupies an infinitesimally small entity such as volume, area or a distance.

The PD functions are constructed such that they are orthogonal to each term in the TSE as

$$\frac{1}{n_1! n_2!} \int_{\mathbf{H}_\mathbf{x}} \xi_1^{n_1} \xi_2^{n_2} g_2^{p_1 p_2}(\boldsymbol{\xi}) dV = \delta_{n_1 p_1} \delta_{n_2 p_2} \quad (A.3)$$

with $(n_1, n_2, p, q = 0, 1, 2)$ and $\delta_{ij}$ is the Kronecker delta symbol. Enforcing the orthogonality conditions in the TSE leads to the nonlocal PD representation of the function itself and its derivatives as

$$f(\mathbf{x}) = \int_{\mathbf{H}_\mathbf{x}} f(\mathbf{x}+\boldsymbol{\xi}) g_2^{00}(\boldsymbol{\xi}) dV \quad (A.4a)$$

$$\begin{Bmatrix} \dfrac{\partial f(\mathbf{x})}{\partial x} \\ \dfrac{\partial f(\mathbf{x})}{\partial y} \end{Bmatrix} = \int\limits_{\mathbf{H_x}} f(\mathbf{x}+\boldsymbol{\xi}) \begin{Bmatrix} g_2^{10}(\boldsymbol{\xi}) \\ g_2^{01}(\boldsymbol{\xi}) \end{Bmatrix} dV \qquad (A.4b)$$

and

$$\begin{Bmatrix} \dfrac{\partial^2 f(\mathbf{x})}{\partial x^2} \\ \dfrac{\partial^2 f(\mathbf{x})}{\partial y^2} \\ \dfrac{\partial^2 f(\mathbf{x})}{\partial x \partial y} \end{Bmatrix} = \int\limits_{\mathbf{H_x}} f(\mathbf{x}+\boldsymbol{\xi}) \begin{Bmatrix} g_2^{20}(\boldsymbol{\xi}) \\ g_2^{02}(\boldsymbol{\xi}) \\ g_2^{11}(\boldsymbol{\xi}) \end{Bmatrix} dV \qquad (A.4c)$$

The PD functions can be constructed as a linear combination of polynomial basis functions

$$g_2^{p_1 p_2} = a_{00}^{p_1 p_2} w_{00}(|\boldsymbol{\xi}|) + a_{10}^{p_1 p_2} w_{10}(|\boldsymbol{\xi}|)\xi_1 + a_{01}^{p_1 p_2} w_{01}(|\boldsymbol{\xi}|)\xi_2 + a_{20}^{p_1 p_2} w_{20}(|\boldsymbol{\xi}|)\xi_1^2 \\ + a_{02}^{p_1 p_2} w_{02}(|\boldsymbol{\xi}|)\xi_2^2 + a_{11}^{p_1 p_2} w_{11}(|\boldsymbol{\xi}|)\xi_1\xi_2, \qquad (A.5)$$

where $a_{q_1 q_2}^{p_1 p_2}$ are the unknown coefficients, $w_{q_1 q_2}(|\boldsymbol{\xi}|)$ are the influence functions, and $\xi_1$ and $\xi_2$ are the components of the vector $\boldsymbol{\xi}$. Assuming $w_{q_1 q_2}(|\boldsymbol{\xi}|) = w(|\boldsymbol{\xi}|)$ and incorporating the PD functions into the orthogonality equation lead to a system of algebraic equations for the determination of the coefficients as

$$\mathbf{A}\mathbf{a} = \mathbf{b}, \qquad (A.6)$$

where

$$\mathbf{A} = \int\limits_{\mathbf{H_x}} w(|\boldsymbol{\xi}|) \begin{bmatrix} 1 & \xi_1 & \xi_2 & \xi_1^2 & \xi_2^2 & \xi_1\xi_2 \\ \xi_1 & \xi_1^2 & \xi_1\xi_2 & \xi_1^3 & \xi_1\xi_2^2 & \xi_1^2\xi_2 \\ \xi_2 & \xi_1\xi_2 & \xi_2^2 & \xi_1^2\xi_2 & \xi_2^3 & \xi_1\xi_2^2 \\ \xi_1^2 & \xi_1^3 & \xi_1^2\xi_2 & \xi_1^4 & \xi_1^2\xi_2^2 & \xi_1^3\xi_2 \\ \xi_2^2 & \xi_1\xi_2^2 & \xi_2^3 & \xi_1^2\xi_2^2 & \xi_2^4 & \xi_1\xi_2^3 \\ \xi_1\xi_2 & \xi_1^2\xi_2 & \xi_1\xi_2^2 & \xi_1^3\xi_2 & \xi_1\xi_2^3 & \xi_1^2\xi_2^2 \end{bmatrix} dV, \qquad (A.7a)$$

$$\mathbf{a} = \begin{bmatrix} a_{00}^{00} & a_{10}^{00} & a_{01}^{00} & a_{20}^{00} & a_{02}^{00} & a_{11}^{00} \\ a_{00}^{10} & a_{10}^{10} & a_{01}^{10} & a_{20}^{10} & a_{02}^{10} & a_{11}^{10} \\ a_{00}^{01} & a_{10}^{01} & a_{01}^{01} & a_{20}^{01} & a_{02}^{01} & a_{11}^{01} \\ a_{00}^{20} & a_{10}^{20} & a_{01}^{20} & a_{20}^{20} & a_{02}^{20} & a_{11}^{20} \\ a_{00}^{02} & a_{10}^{02} & a_{01}^{02} & a_{20}^{02} & a_{02}^{02} & a_{11}^{02} \\ a_{00}^{11} & a_{10}^{11} & a_{01}^{11} & a_{20}^{11} & a_{02}^{11} & a_{11}^{11} \end{bmatrix},$$
(A.7b)

and

$$\mathbf{b} = \begin{bmatrix} 1 & 0 & 0 & 0 & 0 & 0 \\ 0 & 1 & 0 & 0 & 0 & 0 \\ 0 & 0 & 1 & 0 & 0 & 0 \\ 0 & 0 & 0 & 2 & 0 & 0 \\ 0 & 0 & 0 & 0 & 2 & 0 \\ 0 & 0 & 0 & 0 & 0 & 1 \end{bmatrix}.$$
(A.7c)

After determining the coefficients $a_{q_1 q_2}^{p_1 p_2}$ via $\mathbf{a} = \mathbf{A}^{-1}\mathbf{b}$, the PD functions $g_2^{p_1 p_2}(\boldsymbol{\xi})$ can be constructed. The detailed derivations and the associated computer programs can be found in [37]. The PDDO is nonlocal; however, in the limit as the horizon size approaches zero, it recovers the local differentiation as proven by Silling and Lehoucq [42].

**Appendix B - Periodic Activation Function**
Ziyin et al. [36] proposed a new activation function for improving predictive capabilities of a DNN for data with periodic behavior. This activation function is $f_\alpha(x) = x + \frac{1}{\alpha}\sin^2(\alpha x)$, where $\alpha$ is the frequency of the periodic part of the activation function. It is superior for predicting periodic behavior because it has inductive bias for periodicity. Also, it is easier to optimize compared to other periodic activation functions with infinitely many local minima. With this activation function at every layer, they employ a fully connected DNN structure. Ziyin et al. [36] specified the value of parameter $\alpha$ as a fixed parameter. Figure 14 shows the effect of different $\alpha$ values on the behavior of the activation function. They demonstrated the performance of this activation function for image classification, financial, atmospheric and body temperature predictions. It is also capable of predicting time evolution of non-periodic data.

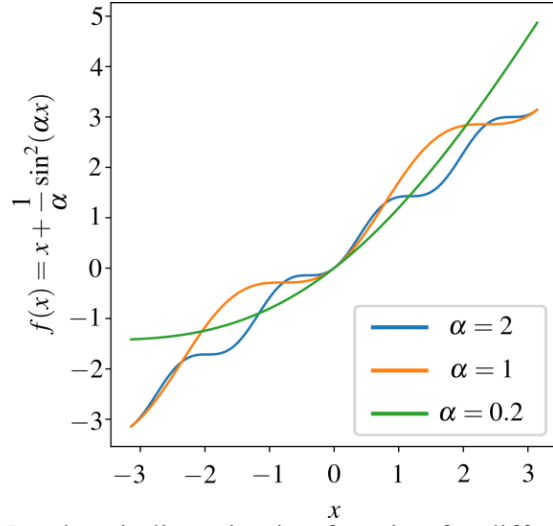

**Figure 14.** Novel periodic activation function for different $\alpha$ values

**Appendix C - Training Losses for Numerical Experiments**

Training losses of the numerical experiments with the present solutions for 2D Coupled Burgers' equation, $\lambda - \omega$ Reaction-Diffusion equation and Gray-Scott equation are shown in Figure 15 as a semilog plot. Models for Burgers' and Gray-Scott equations are terminated before the plateau to prevent the overfitting. The same approach is followed for $\lambda - \omega$ Reaction-Diffusion equation; however, the periodic behavior of the solution creates oscillation in training error. Therefore, it is more difficult to observe the convergence.

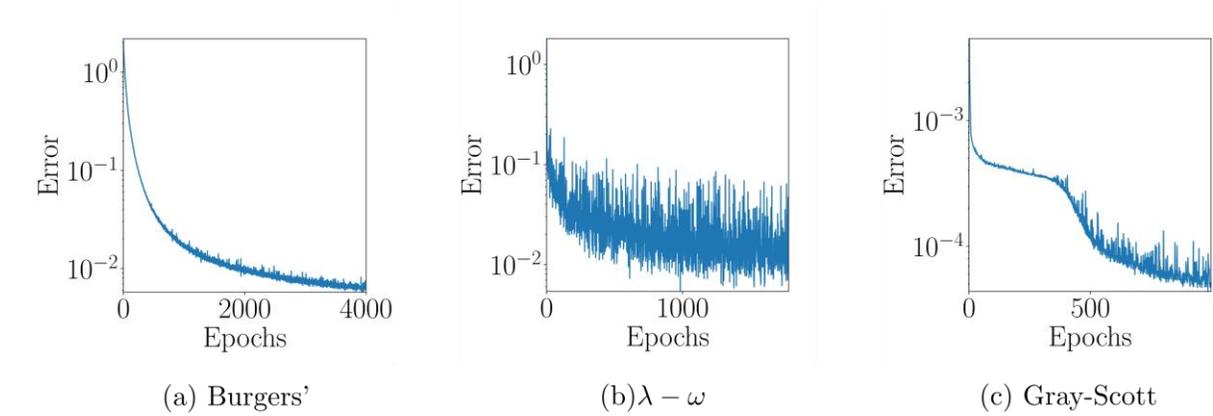

(a) Burgers'  (b) $\lambda - \omega$  (c) Gray-Scott

**Figure 15.** Training histories of the numerical experiments

**Appendix D - Training of PINN Models**

We use SciANN library [43] to build and train the PINN models. Uniform sampling is adopted for the imposition of initial conditions and uniform random sampling for the domain collocation points. Number of collocation points for the initial condition and domain points are specified as $16384(128 \times 128)$ and $327680(20 \times 128 \times 128)$, respectively. Periodic boundary conditions are imposed by using Fourier Features [44]. Our $C^k$ periodic layer, which comes after the input layer, has 20 frequency components. The PINN setup has 4 hidden layers and 80 neurons with tanh activation function while using Glorot uniform initialization [45] for the network weights. Also, we use Adam optimizer [46] with initial learning rate of $10^{-3}$ which exponentially decreases to $10^{-4}$ and loss weights are updated using gradient norm algorithm [22]. We use a batch size of 2000 and trained the PINN for 800 epochs. Training takes 8 hours with (AMD EPYC 7642 16-core) CPU with 2.4 GHz clock speed and 32 GB RAM.